# The Design Space of Social Robots

D.B. Skillicorn, R. Billingsley, M.-A. Williams
Magic Lab, University of Technology Sydney

**Abstract:** We consider the design space available for social robots in terms of a hierarchy of functional definitions: the essential properties in terms of a locus of interaction, autonomy, intelligence, awareness of humans as possessors of mental state, and awareness of humans as social interactors. We also suggest that the emphasis on physical embodiment in some segments of the social robotics community has obscured commonalities with a class of agents that are identical in all other respects. These definitions naturally suggest research issues, directions, and possibilities which we explore. Social robotics also lacks compelling 'killer apps' which we suggest would help focus the community on a research agenda.

> *"A machine capable of carrying out a complex series of actions automatically, especially one programmable by a computer"*

> *"A machine that resembles a human and does mechanical, routine tasks on command"*

> *"Any machine or mechanical device that operates automatically with humanlike skill"*

> *"A machine resembling a human being and able to replicate certain human movements and functions automatically"*

> *"Devices that possesses some of these capabilities: accept electronic programming, process data or physical perceptions electronically, operate autonomously to some degree, move around, operate physical parts of itself or physical processes, sense and manipulate their environment, and exhibit intelligent behavior — especially behavior which mimics humans or other animals"*

As these online dictionary definitions of 'robot' show, there is considerable confusion about what exactly the term 'robot' is supposed to mean. They attempt to blend two quite disparate underlying concepts: a device that perform physical actions, while at the same time resembling a human.

Definitions in the more-formal literature are not much clearer:

> *"A reprogrammable, multifunctional manipulator designed to move material, parts, tools, or specialized devices through various programmed motions for the performance of a variety of tasks"* credited to the Robot Institute of America, 1979.

> *"An automatically controlled, reprogrammable, multipurpose, manipulator programmable in three or more axes, which may be either fixed in place or mobile for use in industrial automation applications"*
> ISO8373 (https://www.iso.org/standard/55890.html)

> *"A robot is a constructed system that displays both physical and mental agency, but is not alive in the biological sense."* (Richards and Smart, 2013)

> *"A social robot is an autonomous or semi-autonomous robot that interacts and communicates with humans by following the behavioral norms expected by the people with whom the robot is intended to interact."* (Bartneck and Forlizzi, 2004)

The Richards and Smart definition, especially, seems to get closer to a working definition of a robot as it is conceived in the research community. However, it is broader than some researchers would agree with because it does not explicitly require embodiment.

In the closest work to what we attempt here, in 2004, Bartneck and Forlizzi attempted to define a design-centered framework for social robots. Their definition of a social robot is given above. They concluded that social robots could be categorised along five dimensions: form (abstract to anthropomorphic), modality (range of communication channels used), social norms (understanding of social expectations), autonomy, and interactivity (ability to respond to human behaviour). From these properties they derived design guidelines which, in the end, are rather weak: e.g. "The form of a social robot should match its abilities". Breazeal et al. (2008) categorise social robots by whether they display paralinguistic information (non-verbal aspects of communication), whether they understand paralinguistic information, whether they include emotion in interactions, and whether they can build and exploit a theory of mind for humans and other robots with which they interact. Hegel et al. (2009) take a semiotic view of social robotics, categorising the design space for robots by form, function, and context, further subdividing function into practical function, aesthetic function, and indicating functions.

Our goal is more ambitious. We suggest, and motivate, definitions of the properties that, first, differentiate robot-like systems from other software/hardware systems and, second, differentiate between less- and more-powerful kinds of robot-like systems. Good definitions need to be precise enough that they accurately characterise the system they reference; when they are precise enough they naturally suggest research directions, which we also explore.

Because of ambiguities about the extent to which the systems we are defining must be embodied, we use the term agent/robot, aware that it is a clunky phrase. We must also contrast what we defining with respect to the more general concept of 'agent' (Poole and Mackworth, 2016). In that community, an agent is anything that has agency, that is which is purposive, possessing goals and/or preferences and means to work towards them. This definition includes living things as well as artificial agents. Furthermore, the agent community defines intelligence in terms of appropriateness for goal seeking, flexibility, and learning. This definition seems too general – a single-celled creature is intelligent according to it and, in software terms, it includes the analytic tool that predicts whether a credit card transaction should be approved or not. The agent community also rules out systems that exist only to think; but this seems perverse. However, the agent community has made great progress with designing and implementing world models, hierarchical planning, and learning, all of which deserve to be more widely used in the robotics community.

Ordinary software/hardware systems have revolutionised the world in the last 70 years. They can be divided into three main categories: those that do not interact with humans directly (for example, network switches), those that interact with humans via information flow (most software), and those that interact with the physical world (for example, Roombas). These distinctions continue to have a role in the world of agent/robots.

What is it that agent/robots can do that such conventional systems cannot? The main answers seem to be that they are easier to direct (because they will understand what we want without detailed instructions); they are general-purpose (so that they will replace the affordances of many current systems, as cell phones have replaced many different portable devices); and they can manipulate the physical world in ways that are beyond current, m ore specialised systems (for example, becoming personal support workers, cleaners, or firefighters). However, one of the missing pieces in robotics research is a clear sense of the 'killer apps' for such systems. Agreement about some set of these would provide a crisper focus for research.

In any research area, good definitions clarify the landscape of what has been done, and the research questions that remain to be answered. The following definitions attempt to capture the key ideas in the design space of 'social robotics'. We begin with the basic differentiator from conventional software/hardware systems.

**Definition**: An agent/robot is a system for which a locus of interaction with the real-world is a critical property.

This definition implies that such a system has to be embodied in at least a virtual sense; the system must 'inhabit' something. Since agent/robots are specifically physically located at any moment in time, they are not interchangeable with one another – they have some form of individuality since a system whose locus of interaction is *here* cannot be the same as one whose locus of interaction is *there*. An agent/robot is therefore a kind of converse of other systems and applications, which are identical to one another, and often available from anywhere. A web server does not have a locus of interaction.

An agent has a locus of interaction within some other device or physical location. The common cases are a locus within a mobile device such as a cell phone, or a locus within a specific room, where it interacts via cameras, microphones, speakers, and screens. An agent has a *mental* embodiment.

A robot has a locus of interaction in a specialised physical construct. This may be fixed (for example, a robotic concierge) or mobile. A robot is embodied in a stronger sense – it has both a mental and a *physical* embodiment.

The key property of an agent/robot is the notion of place: it is a software/hardware systems that operates in a specific 'location', and interacting with it requires that you go to it, or it comes to you. Such systems are not necessarily sophisticated just because they have a locus of interaction. For example, an industrial assembly robot is just a piece of software with an unusual output device – but its outputs only make sense in a specific location. A transit tap-on-tap-off system in a bus is quite a simple agent, but its (current) location is key to its functionality.

For systems closer to what is conceived, at least in the popular mind, as robots extra properties are needed.

**Definition**: An agent/robot is <u>autonomous</u> if, as well as an assigned task list, it has self-generated (including homeostatic) tasks AND it has a regulatory system to manage (order) both kinds of tasks together by internally defined importance.

Of course, any system has the ability to understand commands and, if it is to be useful, to carry out these commands. The new capability of an autonomous system is that it can generate usefulness without, or beyond, commands it has been given, by carrying out tasks that need not be given to it explicitly. An autonomous agent/robot is therefore capable of surprising the humans in its vicinity.

Because such a system has two different kinds of tasks to carry out, new issues arise in deciding what priority to give them; hence the need for a sophisticated regulatory system.

A system whose only self-generated tasks are homeostatic should probably not be considered autonomous. For example, operating systems such as Windows regularly check for updates, and download and install them without user intervention but this is not enough to be considered autonomous. High-performance computational systems such as database query engines monitor their own performance and predict upcoming resource demands, adjusting their configurations, so it is appropriate to consider them as autonomous, but they are not agent/robots.

An autonomous agent has mental agency; an autonomous robot has both mental and physical agency.

We have already indicated that a locus of interaction creates a specific identity for each instance of an agent/robot, since they each interact with the outside world in distinct places. The definition of autonomy strengthens this notion of identity since the actions of an individual system, once it has begun to operate, change its state with respect to the ordering of future actions. An autonomous system is more stateful than a conventional system or, to put it another way, an autonomous system learns in at least a simple sense. Thus systems with identical software and hardware configurations, while already different by being embodied in (a set of) different locations, diverge as the result of the feedback loops between their regulatory system choices, and the outcomes of the chosen actions.

Therefore, hard rebooting an autonomous agent/robot (i.e. without state preservation) produces a new instance.

There are very few examples of autonomous systems in this sense in existence, and they tend to have been designed to operate in constrained environments where the task possibilities are limited, and so the regulation process is manageable. There is also perhaps a pragmatic difficulty: conventional software can be replicated with low marginal cost, but this is less true for an autonomous agent/robot.

**Definition**: An autonomous agent/robot is <u>intelligent</u> if its regulatory system prevents it from getting stuck.

This definition implies that, as well as techniques such as generalized timeouts, the system predicts the consequences of its actions and uses this to determine their scheduling (endo-prediction). Thus it is capable of, for example, determining the energy cost of an action and not carrying it out if to do so would leave it unable to recharge.

This idea of not getting stuck is not intended to (somehow) evade the limits on computability. As long as problematic tasks do not necessarily have to be completed, mechanisms such as timeouts can avoid the pathological computational cases.

Many other definitions of intelligence are based on goal seeking behaviour, but living intelligent things seek their *own* goals. Agent/robots, on the other hand, are tools and so their goals derive from the needs and desires of humans (and only then their own existence). This apparently small distinction means that much of the theory of intelligence does not directly apply to agent/robot intelligence. (Asimov's famous Three Laws of Robotics captured the primacy of human interests over robot interests, but failed to take into account the ownership of robots, a point we return to below.)

**Definition**: An intelligent agent/robot is <u>social</u> if it can infer tasks and their importance based on updatable models of the humans around it.

This definition requires that an agent/robot has a set of models of humans it may or has encountered, is proactive in refining and updating these models of humans based on what it sees and infers, and does exo-prediction of what these humans want/will do, based on the models. The management and updating of these models becomes a major component of its self-generated tasks.

This definition allows a robot that navigates along a busy pavement to be considered a social robot, even if it doesn't interact explicitly or directly with any human, since it must have models of how humans move in order to be able to thread its way through them; it will perform better if it can infer from the appearance of each human how he or she is likely to move in the immediate future; and it will also perform better if it is aware of social conventions such as keeping to the right or left. The comparison with self-driving cars is instructive. At present, such cars model pedestrians in essentially the same way that they model other vehicles, as moving objects. A social self-driving car would be able to infer from the movement of a particular pedestrian that he is drunk, and would build this into predictions of his future movements.

It is also necessary for a robot to have models of other robots since: it will encounter them physically (and so must be aware that they, unlike other moving objects, have agency); it must collaborate with them to divide up tasks in many plausible settings; and it can learn and teach by sharing knowledge and models with them.

Robots must also have models of other beings with agency, i.e. animals, whose future behaviour cannot be fully modelled by information about their recent behaviour.

This definition of social agent/robot allows for a further, more powerful class of social agent/robots, which might perhaps be called *verbal social robots* or *interacting social robots*, which have the capability of communicating with humans directly. Breazeal (2002) defined sociable robots that are quite close to these but required also that they should be lifelike in order to leverage human tendencies to anthropomorphism. Sometimes this argument is made along the lines of the need to signal human-like affordances via a human-like appearance (Hegel et al. 2009). However, an agent like HAL in the *2001: A Space Odyssey* film seems clearly to be an interacting social agent (albeit an evil one) despite the lack of a physical body. Humans manage to interact socially with other humans even when disembodied, using the telephone; and with limited embodiment, using tools like Skype. It could be argued that humans understand the sociality of characters in novels, with no embodiment at all. Thus, the necessity of embodiment for social agent/robots is perhaps overestimated.

Obviously, the ability of an agent/robot to interact with humans using the full range of modalities humans use with each other greatly increases the sophistication needed for the interface, and also the richness of the models of humans that the robot must construct.

There is a natural hierarchy of capability in agent/robot systems: locus of interaction, autonomy, intelligence, social modelling, and social communication.

We can also differentiate agent/robots along axes such as: learning (can they acquire new facts, or can they also infer new models, or can they also alter their own regulatory system); and communication sophistication (must they be instructed directly – programmed – or can they also understand imprecise or implicit commands; or can they also infer human desires from their models of humans, without any communication) (Billingsley et el. 2017).

The way we have framed these definitions implies that the software/systems part of a robot is necessarily coupled with its physical manifestation. It is conceivable that the 'soft' part of a robot could migrate to another physical manifestation. This would avoid the need for a robot to go from one place to another just because its associated human did. For example, shopping centers could provide robot 'bodies' that could be occupied by the 'soft' part of personal support workers accompanying shoppers. However, there are substantial practical difficulties with this idea – incompatibilities between the hardware configurations are likely to create bumpy experiences of the kind known to those who try to use someone else's computer (or even a different keyboard).

## Examples.

It is helpful to test these definitions in the space of existing, and readily conceived, systems to see how they should be categorised. We illustrate this in the following table. Systems in italics do not exist today, but seem reachable within a reasonable time frame.

| Characteristic | Interaction | AGENTlike | | ROBOTlike | |
|---|---|---|---|---|---|
| | | fixed | carried | fixed | mobile |
| Locus of interaction | no human interaction | network switches | | industrial assembly | automatic train; robot firefighter |
| | 1-1 | | Siri etc.; augmented reality glasses; *paraplegic exoskeleton* | | |
| +Autonomous | no human interaction | | | | UAV (in loiter mode); Mars Rovers; bomb disposal robot |
| | 1-1 | | | | UAV (in piloted mode) |
| +Intelligent | no interaction | | | | pipeline cleaner; warehouse management robot |
| | 1-1 | | | | self-driving car (passengers); |
| | 1-many | | | | self-driving car (pedestrians) |
| +Social | 1-1 | *smart information kiosk; teacher* | *augmented reality glasses* | *airport check-in luggage handler* | *robot pet; home help; porter* |
| | 1-many | *roombot (e.g. HAL)* | *personal lawyer-bot* | *smart triage* | health/aged interactions *chaperone; security and safety; delivery-bot* |

An important differentiator is whether the agent/robot operates without human interaction, interacts with a single human (and so is primarily a physical, mental, or social enhancement for its user), or interacts with multiple humans simultaneously, and so is truly social.

There are many empty squares in the table. Exemplars in some of these can be imagined – for example, an intelligent agent could patrol the Internet looking for malware and removing it. This requires intelligence so that it doesn't get stuck, computationally and geographically, but does not require interaction with humans. It is striking, though, how much both mobility and social abilities increase the space of interesting potential exemplars.

Systems that have a locus of interaction but no other sophisticated properties are the totality of those that are considered as robots from a lay perspective, but they are really the simplest agent/robot systems.

Autonomous systems are being developed rapidly, although more within an agent framework than a robot framework. Such systems would naturally be improved by adding intelligence. Today, the most sophisticated systems that are intelligent function within frameworks that are limited physically. For example, unmanned aerial vehicles (UAVs) can loiter over a particular area for long periods of time, requesting help from a human pilot only when something interesting happens.

It is difficult to assess the functionality of existing social agent/robots. Many presentations of the functioning of such systems are scripted; when the results of experiments are reported, they are either of limited scope, or scored in opaque ways, or both.

As the table shows, there are abundant opportunities for social robots to perform useful roles. The capabilities required to be social are largely orthogonal to the capabilities required by physical embodiment, except that intelligence interacts with mobility because of the need to manage power.

## Research Directions

These definitions suggest at least the following research directions:

General:

Building robots requires considerable expense and complexity. Many of the research questions being addressed can be done as well within an agent framework. Some examples where progress could be made (even though not every question could be answered) include: learning from humans and teaching them, collaborative planning, engagement, detecting and transmitting emotions and moods, and detecting mental states in humans and simulating them in an agent. Much can be adapted from the wider software agent community, but issues of localization, embodiment, and close interactions with humans create novel issues.

Conversely, many of the research questions of mobility and embodiment being addressed can be done without the need for much of the social framework. Some examples include: path and motion planning, optimal telepresence, reference resolution, and haptic interaction.

Of course, movement is a key aspect of human sociality, so that these two directions will eventually overlap. But perhaps more progress could be made with a reductionist approach.

Related to autonomy:

Regulatory systems that understand the difference between commanded tasks and self-generated tasks, and can make decisions about the order in which to complete them, are needed. It seems necessary to consider a regulatory system as a hierarchy of regulatory systems acting at different levels of abstraction. At the top level, the regulatory system decides (from moment to moment) the ranking of a set of goals. Each goal consists of a set of tasks; at the next level, the regulatory system ranks the available tasks. Each task consists of a set of actions; at the next level the regulatory system ranks the actions, and chooses the one to execute. At any time, new input from the external environment and the internal state of the robot can change the objects in play and their rankings at any or all of these levels of abstraction. For example, in humans reflexes re-order actions at a very low level, and without necessarily re-ordering higher level tasks, while other physiological signals seem to operate by re-ordering tasks or even goals; for example, getting hungry. Robots will need to have similar flexibility. Algorithmic approaches to choosing what to do next (maximising utility) are typically not framed in this hierarchical way at present, although the concept of hierarchical control is an old one (Albus, 1991). For example, the popular ROS system provides a single shared space in which both publish-subscribe and remote procedure calls operate. Again there is much to be learned from the software agent community, but new issues are in play for social agent/robots.

What repertoire of goals, actions, and tasks are available to an agent/robot, and where do they come from? Can the agent/robot create effectively new ones by combining existing ones, or can it infer genuinely new ones (i.e. a substantive form of learning)?

Who has the right to command an agent/robot to acquire a goal, or perform a task, and how are conflicts between competing commands to be resolved? Must self-generated tasks always be lower priority than commanded tasks, or may an agent/robot refuse to obey a command because of its own 'needs' (e.g. charging)? Since agent/robots will be expensive for some time, especially as their individuality means high marginal costs for each unit, they will have owners who expect the agent/robots to serve their specific goals, certainly with higher priorities than requests from random human passers-by.

How can robots agree amongst themselves which is to carry out any particular task in a shared context? In human societies, this usually results in some form of specialisation – one individual tends to complete particular tasks, even though all are equally capable. In an environment where a skill can be encoded and transmitted from one robot to another, does specialisation have a role to play? Will agent/robots be mass-produced, or will particular constraints mean that they will be highly variable (as cars are today)?

Related to intelligence:

If a system is not to get stuck, it must either know before it starts an activity that the activity can be completed in a timely way, or it must have a mechanism for bailing out of the activity when it becomes clear that it cannot be completed (possibly reconsidering at some higher level and starting a new task to achieve the same end). Therefore, intelligence requires a system to predict what will happen when any particular goal, task, or action is performed, and a way to compare what is actually happening with what was predicted. The simplest form of such a system would predict how long each activity should take, and timeout if it has not completed. In humans, the cerebellum is responsible for predicting what is supposed to happen as the human moves; a similar mechanism seems essential for robot movement; and something similar happens in humans as we follow a planned activity. For a robot, it is also essential to predict how much power the activity will consume to ensure that it can be done within the context of current available power.

Although intelligent systems need not be social, an awareness of human timeframes, combined with the ability to predict the time required for a task, improves usability. In current systems, it is easy for a user to begin a computation that takes far longer than anticipated; flagging this in advance would prevent wasted work and time. (Of course, prediction of required time is not possible in the general case, but is for most practical purposes.) Software that takes 1s, 10s, 100s, 1000s (=16 min), 10,000s (=3 hours), or 100,000s (=30 hours) cause quite different interaction behaviour with humans. It would be useful to expose the system's understanding of estimated time for tasks to users.

Related to social interaction:

The key requirement is to be able to build models of humans that are rich enough to have an effective impact on goal, task, and action generation and ranking. Such models can vary in complexity. The simplest might be a movement model that predicts velocity from visual factors such as height and weight; a more complex model might recognise faces and so remember which humans have been met before; a still more complex model might estimate properties of individuals such as interests, roles, attitudes, and personality. The goal of research is to allow the agent/robot to build models of humans that are similar to those that humans build of other humans. As an extension, robots also need to build models of other robots and animals, with the same range of possible complexities.

The second research challenge is to define and build the techniques for enabling these models of humans to influence the agent/robot's goals, tasks, and actions. Conceptually this requires the agent/robot to be able to perceive the goal-task-action priorities of humans and compute the feedback loops between them and its own goal-task-action priorities. At the simplest level this requires not running into or hitting an adjacent human who is also moving. At its most sophisticated, this requires implementing the general concept of 'being helpful', that is, assisting the human to achieve his or her goals.

A third research challenge is to define a legal system for humans and robots to interact. Many of the problems associated with self-driving cars also apply to social robots: who is responsible when a robot injures a human? Can robots (in security or care settings) force humans to do things they don't want to? Must robots always be truthful? If a corporation can be a legal person, can a robot be one too? Although Asimov explored the implications of his Three Laws extensively, there remain many difficult edge cases, and he did not allow full force to ownership, and by implication control, of a robot.

Research in social robotics is surprisingly fragmented, with little sense of the roadmap of problems that must be solved, and their necessary sequencing. The contribution of this paper is to suggest both problems that seem to be understudied and potentially fruitful directions for further work.

## References

J. Albus, Outline for a Theory of Intelligence, IEEE Transactions on Systems, Man, and Cybernetics, Vol. 21, No. 3, May-June 1991, 473-509.

C. Bartneck and J. Forlizzi, A Design-Centred Framework for Social Human-Robot Interaction, Proceedings of 2004 IEEE International Workshop on Robot and Human Interactive Communication, Japan, 2004, 591-594.

R. Billingsley, John Billingsley, P Gärdenfors, P. Peppas, H. Prade, D. Skillicorn and M-A. Williams, The Altruistic Robot: do what I want, not just what I say, Scalable Uncertainty Management (SUM) 2017, Springer, 149-162.

C. Breazeal, Designing Sociable Robots, MIT Press, 2002.

C. Breazeal, A. Takanishi, and T. Kobayashi, Social Robots that Interact with People, Chapter 58, Springer Handbook of Robotics, 1349-1369, 2008.

F. Hegel, C. Muhl, B. Wrede, M. Hielscher-Fastabend, G. Sagerer, Understanding Social Robots, Second International Conferences on Advances in Computer-Human Interactions, 2009, 169-174.

David L. Poole and Alan K. Mackworth, Artificial Intelligence: Foundations of Computational Agents, Cambridge University Press, 2nd Edition, 2016.

N.M. Richards and W.D. Smart, How should the law think about robots? Available at SSRN: https://ssrn.com/abstract=2263363, May 10, 2013.